\documentclass[journal]{IEEEtran}

\usepackage{cite}
\usepackage{amsmath,amssymb,amsfonts}
\usepackage{amsthm}
\usepackage{graphicx}
\usepackage{booktabs}
\usepackage{multirow}
\usepackage{algorithm}
\usepackage{algorithmic}
\usepackage{textcomp}
\usepackage{xcolor}
\usepackage{url}
\usepackage{stfloats}
\usepackage{placeins}
\usepackage{siunitx}
\usepackage{mathrsfs}
\usepackage{hyperref}
\hypersetup{hidelinks}

\begin{document}

\title{Efficient Tuning Before Low-Bit Post-Training Quantization for Stochastic Gradient Descent-optimized Models}

\author{
Peng~Xia,
Junbiao~Pang*,
Muhammad Ayub Sabir
\thanks
{
\IEEEcompsocthanksitem * Corresponding author(s).

\IEEEcompsocthanksitem Peng Xia, Junbiao Pang, and M. Sabir are with the School of Information Science and Technology, Beijing University of Technology, Beijing 100124, China (e-mail: \mbox{junbiao\_pang@bjut.edu.cn})
}
}

\maketitle

\begin{abstract}
Post-training quantization (PTQ) compresses deep neural networks for deployment under limited memory and computational budgets. However, low-bit (i.e., 2-bit or 4-bit) PTQ often suffers from substantial performance degradation. Most existing PTQ methods operate on an unconstrained full-precision (FP) model and primarily address quantization errors through post-hoc reconstruction. We argue that low-bit PTQ accuracy is limited not only by post-quantization error minimization, but also by the quantization-error tolerance of a FP model itself. In this paper, we propose Efficient Tuning Before Quantization (ETBQ), a pre-conditioning tuning stage for Stochastic Gradient Descent (SGD)-optimized models before PTQ. During tuning, the FP model is optimized under perturbations sampled from the error distributions of weight and activation quantization, guiding the model toward a loss-landscape region that is less sensitive to the subsequent PTQ. Unlike QAT, ETBQ does not train a fake-quantized deployment model, which is computationally and memory intensive. Instead, ETBQ outputs a FP model that can be used by any PTQ backend. Experiments on CIFAR-100, Tiny-ImageNet, ImageNet, and Cityscapes provide consistent evidence that ETBQ improves low-bit PTQ across diverse tasks. Under W2A4 settings, e.g., ETBQ improves over naive PTQ by 2.14\% top-1 accuracy on Tiny-ImageNet and by 5.80\% mIoU on Cityscapes. Code is available at~\url{https://github.com/xpxpxp2001xpxpxp/ETBQ}.
\end{abstract}


\section{Introduction}

\IEEEPARstart{T}{he} deployment of Deep Neural Networks (DNNs) on edge devices has made model compression a critical challenge in efficient deep learning. Among compression techniques, model quantization~\cite{librecq,esser2020learned} is particularly attractive, as it reduces memory usage and enables low-precision integer computation. Post-Training Quantization (PTQ)~\cite{nagel-adaround-icml-2020,weiqdrop,liu-flatquant-iclr-2025} is an efficient paradigm that avoids full-dataset retraining required by Quantization-Aware Training (QAT)~\cite{bhalgat-lsq+-cvpr-2020,shin-nipq-cvpr-2023,fqat-acmmm-2025}. Despite its efficiency, PTQ suffers significant accuracy degradation at low-bit widths (i.e., 2-bit or 4-bit) when both weights and activations are quantized.

\begin{figure}[t!]
    \centering
    \includegraphics[width=1.0\linewidth]{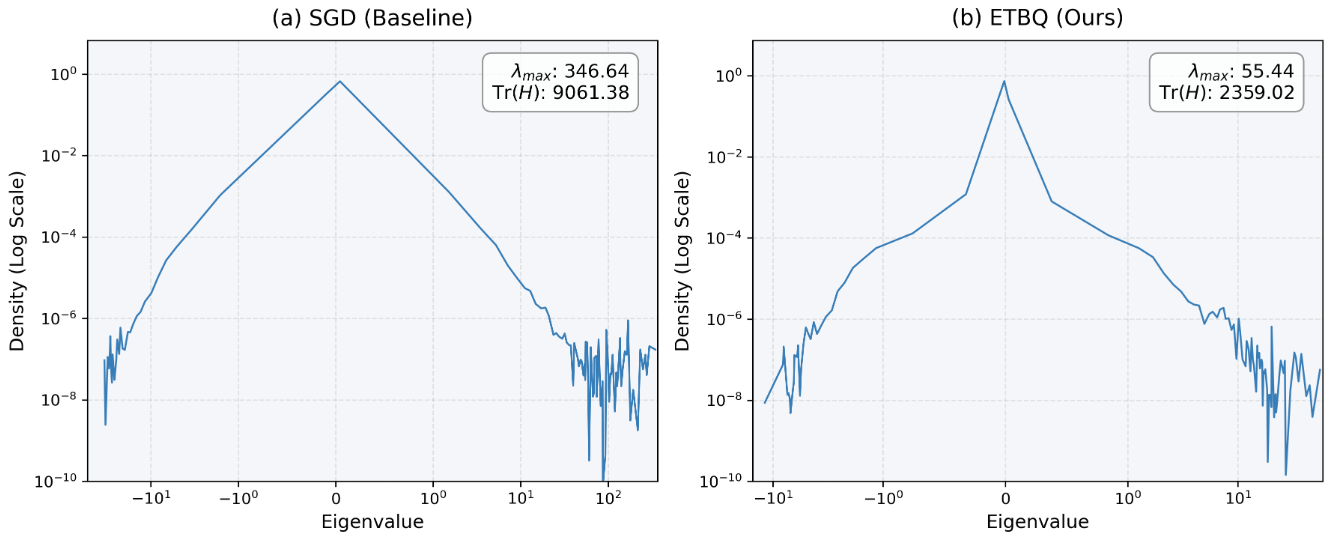}
    \caption{Comparison between Hessian spectral densities of ResNet-18 on CIFAR-100. (a) Standard SGD model, with W2A4 accuracy of 76.41\% (b) ETBQ-preconditioned model, with W2A4 accuracy of 78.06\%. ETBQ shifts the Hessian spectrum toward smaller eigenvalues, indicating a flatter loss basin and improved tolerance to low-bit quantization errors.}
    \label{fig:figure1}
\end{figure}

Existing PTQ methods mainly address accuracy degradation by mimicking the behavior of FP models. Representative approaches include learned weight rounding~\cite{nagel-adaround-icml-2020}, block-wise output reconstruction~\cite{hubaraimproving,librecq}, stochastic activation perturbation~\cite{weiqdrop}, and prediction-difference-guided calibration~\cite{liu2023cvpr-pdquant}. More recent extensions sequentially reduce activation and weight errors~\cite{10.5555/3692070.3694620} or recover quantization-induced information via lightweight compensation modules~\cite{fu2025quantization}. A fundamental question overlooked by conventional PTQ paradigms is: \textit{What intrinsic properties of FP models make them quantization-friendly for low-bit PTQ?} An intuitive yet largely underexplored hypothesis is that FP models must be robust to quantization-induced errors. 

We hypothesize that FP models residing in flatter loss basins are more likely to be quantization-friendly. To examine this hypothesis, we compare a standard SGD-optimized ResNet-18 with its ETBQ-preconditioned counterpart on CIFAR-100. As shown in Fig.~\ref{fig:figure1}(a), the standard SGD baseline has a largest Hessian eigenvalue of $\lambda_{\max}=346.64$ and a Hessian trace of $\operatorname{Tr}(H)=9061.38$, and obtains 76.41\% W2A4 accuracy after QDrop~\cite{weiqdrop}. After ETBQ pre-conditioning, Fig.~\ref{fig:figure1}(b) shows that these two curvature indicators decrease to $\lambda_{\max}=55.44$ and $\operatorname{Tr}(H)=2359.02$, while the W2A4 accuracy improves to 78.06\%. Therefore, a smaller $\lambda_{\max}$ implies a smaller worst-case loss increase under bounded quantization errors, while a smaller Hessian trace indicates lower aggregate curvature across directions. In this sense, ETBQ  moves the FP model to a flat region where the same weight- and activation-side quantization errors cause less loss degradation. These empirical findings establish a correlation, rather than a strict causal proof, between flatter FP loss basins and improved tolerance to quantization errors, supporting ETBQ as a complementary direction to existing PTQ calibration techniques.

In this work, we introduce ETBQ, a dedicated pre-conditioning step performed prior to PTQ to improve the FP model's tolerance to quantization error. ETBQ fine-tunes the SGD-trained FP model by injecting statistical quantization perturbation on weights and activations, then applies Stochastic Weight Averaging (SWA) to stabilize and consolidate the fine-tuning trajectory. The resulting model remains in full precision and can be directly fed into any reconstruction-based PTQ backends without any modification. Our main contributions are summarized as follows:
\begin{itemize}
\item To the best of our knowledge, we introduce a pre-quantization robustness perspective for efficiently tuning quantization-friendly FP models before low-bit PTQ. ETBQ requires access to the training data for an additional FP pre-conditioning stage, and is therefore positioned between calibration-only PTQ and full QAT in terms of data and compute cost.
\item We present the ETBQ framework to mitigate the sensitivity of FP models to quantization errors. Specifically, Weight Quantization Noise (WQN) models weight quantization error and employs temporal differencing to control drift and encourage flatter solutions; meanwhile, Activation Quantization Noise (AQN) models activation quantization error and applies a salience-aware stochastic mask to improve activation-side perturbation robustness.
\end{itemize}

\section{Related Work}
\subsection{Post-Training Quantization}

PTQ compresses a FP model into a low-bit model with a small calibration set, avoiding the resource-intensive retraining process used in QAT. Early data-free methods, such as DFQ~\cite{nagel-dfq-iccv-2019} and ZeroQ~\cite{cai2020zeroq}, improve quantizer initialization through weight equalization or synthetic calibration data. Reconstruction-based methods further reduce quantization error by optimizing rounding decisions or output reconstruction with a small held-out set. For example, AdaRound~\cite{nagel-adaround-icml-2020} introduces a learnable weight-rounding scheme using second-order information. AdaQuant~\cite{hubaraimproving} jointly optimizes weights and quantization parameters via layer-wise reconstruction. BRECQ~\cite{librecq} extends this to block-wise reconstruction using the Fisher Information Matrix. QDrop~\cite{weiqdrop} introduces stochastic activation dropping during reconstruction to improve robustness against activation quantization. MRECG~\cite{ma-mrecg-cvpr-2023} adaptively selects reconstruction granularity to mitigate oscillation. Recent methods further refine the calibration objective: PD-Quant~\cite{liu2023cvpr-pdquant} determines quantization parameters using prediction-difference metrics, ERQ~\cite{10.5555/3692070.3694620} sequentially reduces activation and weight quantization errors, and QwT~\cite{fu2025quantization} compensates quantization-induced information loss through lightweight auxiliary modules. QEP~\cite{arai2026quantizationerrorpropagationrevisiting} observes that layer-wise PTQ errors compound along the network and reformulates the per-layer objective to explicitly propagate and compensate for the accumulated error, with a tunable propagation strength that balances compensation against calibration overfitting.

Beyond CNNs, PTQ4ViT~\cite{yuan-ptq4vit-eccv-2022}, FQ-ViT~\cite{lin-fqvit-ijcai-2022}, and RepQ-ViT~\cite{li-repqvit-iccv-2023} address the distributional challenges in vision transformers, while FlatQuant~\cite{liu-flatquant-iclr-2025} learns per-layer affine transformations that suppress outliers in the weights and activations of large language models, flattening their value distributions so that fewer extreme values stretch the quantization range. This notion of ``flat'' is different from the loss-landscape flatness studied in this paper: FlatQuant reshapes tensor value distributions for easier clipping and scaling, whereas ETBQ aims to move the FP model toward a parameter region whose task loss is less sensitive to quantization-induced perturbations.

\begin{table*}[t]
\centering
\caption{Comparisons among calibration-only PTQ, QAT, and ETBQ. MobileNetV2 has about 3.5M FP parameters; ImageNet has about 1.28M training images, and the calibration subset contains 1024 images in our protocol.}
\label{tab:method_position}
\renewcommand{\arraystretch}{1.15}
\setlength{\tabcolsep}{5pt}
\begin{tabular}{lccc}
\toprule
\textbf{Aspect} & \textbf{Calibration-only PTQ} & \textbf{QAT} & \textbf{ETBQ} \\
\midrule
Optimization stage
& After FP training
& During quantized-model training
& Before PTQ, after FP training \\

Optimized object
& Quantizer / reconstruction objective
& Fake-quantized deployment model
& FP model conditioned for later PTQ \\

Use of full training set
& No
& Yes, full ImageNet training
& Yes, full ImageNet pre-conditioning \\

Calibration subset
& 1024 images for PTQ
& Optional for observer initialization
& 1024 images for error statistics and PTQ \\

Updated FP parameters
& 0
& $\sim$3.5M
& $\sim$3.5M \\

Quantization parameters
& Optimized during PTQ
& Jointly optimized with weights
& Not optimized during pre-conditioning \\

Forward/backward path
& Calibration / reconstruction only
& Fake-quantized forward and backward
& FP forward and backward with quantization errors \\

Typical full-data schedule
& None
& 200 epochs in full-QAT reference
& 80 epochs in our pre-conditioning stage \\

Output before deployment
& Quantized model
& Quantized model
& Pre-conditioned FP model for any PTQ backend \\

Paradigm role
& Low-cost post-hoc calibration
& End-to-end quantized training
& Upstream FP tuning before PTQ \\
\bottomrule
\end{tabular}
\end{table*}

\subsection{Flatness and Quantization Robustness}

Flatness of the loss landscape is closely linked to model generalization. For instance, SWA~\cite{DBLP:journals/corr/abs-1803-05407} averages weights along the SGD trajectory to find wide basin solutions. SAM~\cite{foret2021sharpnessawareminimizationefficientlyimproving} explicitly optimizes worst-case perturbed loss. Anti-PGD~\cite{orvieto2022anticorrelated} connects perturbed gradient evaluation with Hessian-trace-related smoothing. Flatness has been adapted for quantization. For example, QDrop~\cite{weiqdrop} and Bit-Shrinking~\cite{lin-bitshrinking-cvpr-2023} improve flatness during PTQ reconstruction. NIPQ~\cite{shin-nipq-cvpr-2023} replaces the straight-through estimator with pseudo-quantization perturbation, which both stabilizes gradient estimation and implicitly smooths the loss surface. FQAT~\cite{fqat-acmmm-2025} perturbs intermediate features during QAT and adds a feature-distillation term, which can be shown to implicitly regularize the Hessian and thereby flatten the landscape of the quantized model. ETBQ keeps the trainable model in FP, never optimizes a quantization parameter during pre-conditioning. It is therefore best understood as a PTQ-upstream FP pre-conditioning method with higher cost than calibration-only PTQ but lower cost than full QAT (see Appendix~\ref{app:additional_ablation}).

\textbf{Position of ETBQ.}
PTQ, QAT, and ETBQ differ mainly in the stage at which optimization is performed.
PTQ fixes the FP model and performs calibration or reconstruction after the
quantizer is specified. QAT trains a fake-quantized deployment model and jointly
updates weights and quantization parameters. ETBQ instead performs an upstream
FP pre-conditioning stage: it updates the FP weights using quantization-error
statistics, but leaves quantizer optimization to the downstream PTQ backend.

\section{Preliminaries}
\label{sec:method_prelim}

\textbf{Basic Notations.} Let $f(\cdot;\boldsymbol{W})$ denote a FP model with trainable weights $\boldsymbol{W}$, and let $f(\cdot; \hat{\boldsymbol{W}}, \boldsymbol{s}, \boldsymbol{z})$ denote its quantized counterpart parameterized by quantized weights $\hat{\boldsymbol{W}}$, step size $\boldsymbol{s}$, and zero-point $\boldsymbol{z}$. Without loss of generality, given a generic linear projection layer (fully connected or convolutional), let $\boldsymbol{W} \in \mathbb{R}^{D_\text{out} \times D_\text{in}}$ denote the weight matrix, and $\boldsymbol{a}_l \in \mathbb{R}^{D_\text{in}}$ denote the input activation to layer $l$. The model input is denoted $\boldsymbol{x}_0 \in \mathscr{D}_t$, drawn from the training set $\mathscr{D}_t$, and a small subset $\mathscr{D}_c \subset \mathscr{D}_t$ is reserved for PTQ calibration.

\textbf{Quantization.} We use channel-wise quantization for weights and per-tensor quantization for activations, except for post-Softmax layers. The uniform quantization operator maps an FP tensor $\boldsymbol{u}$ (i.e., weight $\boldsymbol{W}$ or activation $\boldsymbol{a}_l$) to integer $\boldsymbol{u}_\text{int}$ and dequantized values $\hat{\boldsymbol{u}}$ as
\begin{equation}
\label{eq:quantization}
\begin{aligned}
    \boldsymbol{u}_\text{int} &= \operatorname{clip}\!\Big(\big\lfloor \frac{\boldsymbol{u}}{\boldsymbol{s}} \rceil + \boldsymbol{z},\, 0,\, 2^q-1\Big),\\
    \hat{\boldsymbol{u}} &= (\boldsymbol{u}_\text{int} - \boldsymbol{z}) \cdot \boldsymbol{s},
\end{aligned}
\end{equation}
where $\lfloor \cdot \rceil$ denotes rounding to the nearest integer, and $q$ is the target bit-width. The scale $\boldsymbol{s}$ and zero-point $\boldsymbol{z}$ are initialized from the calibration subset $\mathscr{D}_c$:

\begin{align}
    \boldsymbol{s} &= \frac{\boldsymbol{u}_\text{max} - \boldsymbol{u}_\text{min}}{2^q - 1}, \quad
    \boldsymbol{z} = \operatorname{clip}\Big(\big\lfloor q_\text{max} - \frac{\boldsymbol{u}_\text{max}}{\boldsymbol{s}} \rceil,\, 0,\, 2^q-1\Big),
\end{align}
where $q_\text{max}=2^q-1$ and $\boldsymbol{u}_\text{max}, \boldsymbol{u}_\text{min}$ are computed over the calibration set.

\textbf{Quantization Error and Objective.} The quantization errors for weights and activations are defined as $\delta \boldsymbol{W} = \hat{\boldsymbol{W}} - \boldsymbol{W}$ and $\delta \boldsymbol{a}_l = \hat{\boldsymbol{a}}_l - \boldsymbol{a}_l$, respectively, and the layer-wise reconstruction error is measured by the mean squared error (MSE) between FP and quantized outputs:

\begin{equation}
\label{eqt:quantization_delta}
    \begin{aligned}
    \mathcal{L}_{\mathrm{MSE}} &= \mathbb{E}_{\boldsymbol{a}_l} \left[ \|\boldsymbol{W}\boldsymbol{a}_l - \hat{\boldsymbol{W}}\hat{\boldsymbol{a}}_l\|^2_2 \right] \\
    &= \mathbb{E}_{\boldsymbol{a}_l} \left[ \|\boldsymbol{W}\boldsymbol{a}_l - \left(\boldsymbol{W} + \delta \boldsymbol{W} \right)\left(\boldsymbol{a}_l + \delta \boldsymbol{a}_l \right)\|^2_2 \right] \\
    &= \mathbb{E}_{\boldsymbol{a}_l} \Big[ \|\underbrace{\boldsymbol{W}\delta \boldsymbol{a}_l}_{\text{AQE}} 
    + \underbrace{\delta \boldsymbol{W} \boldsymbol{a}_l}_{\text{WQE}} 
    + \underbrace{\delta \boldsymbol{W} \delta \boldsymbol{a}_l}_{\text{Second-Order}}\|^2_2 \Big].
    \end{aligned}
\end{equation}

Equation~\eqref{eqt:quantization_delta} reveals that the reconstruction error arises from both Activation Quantization Error (AQE) and Weight Quantization Error (WQE). ETBQ aims to render the FP model robust to both error types from an optimization perspective. 

\textbf{Transferring activation quantization error or jointly optimizing both weight and activation errors?} Equation~\eqref{eqt:quantization_delta} shows that low-bit quantization introduces two error sources: the weight quantization error $\delta\boldsymbol{W}$ and the activation quantization error $\delta\boldsymbol{a}_l$. These two errors can be handled in two different ways: 1) One strategy is to transfer the activation-side error into an equivalent weight-side error, so that both effects are optimized through a single weight-side objective. 2) The second is to condition the FP model jointly on weight and activation errors while keeping each error in its native space. ETBQ adopts the latter design. 

\begin{figure}[h!]
    \centering
    \includegraphics[width=1.0\linewidth]{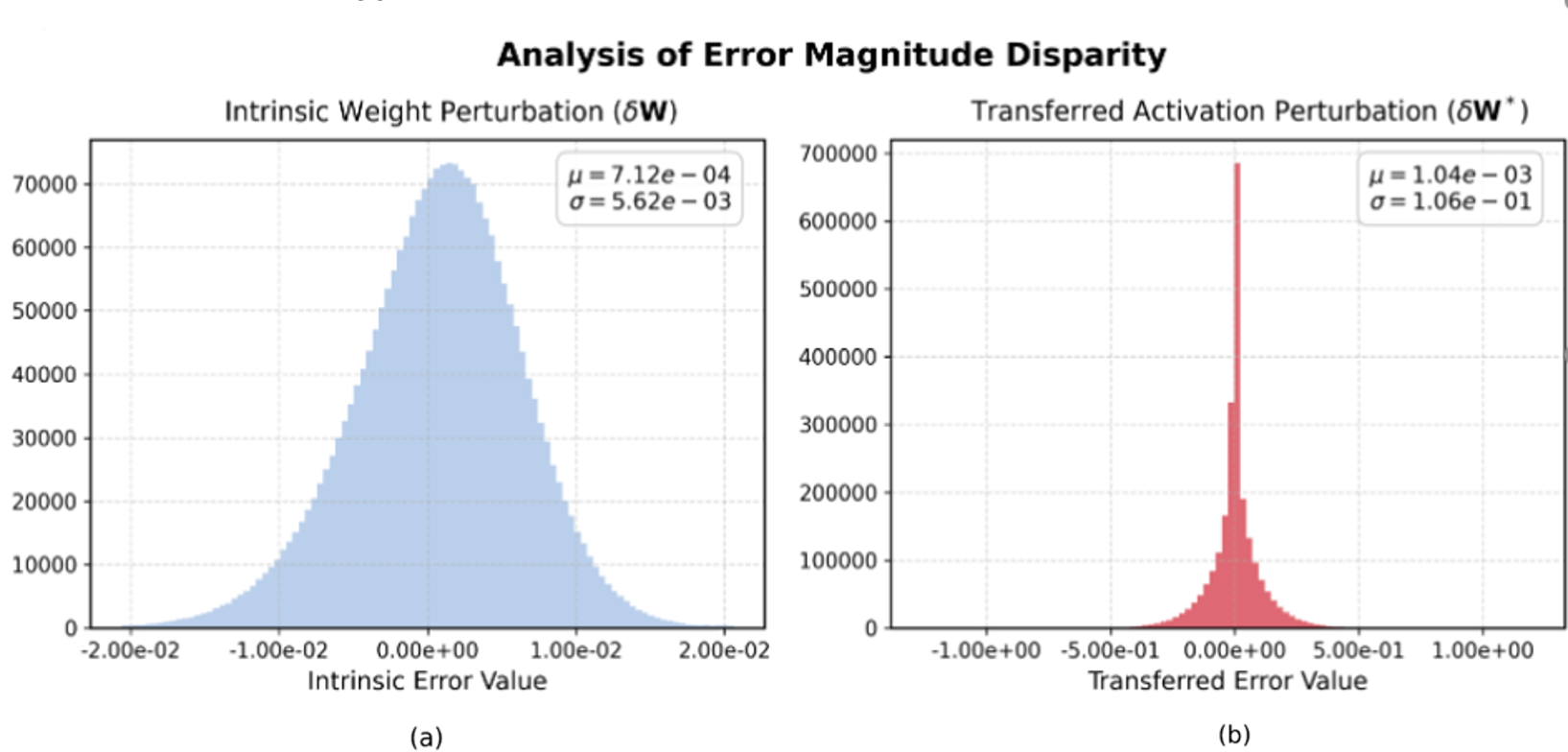} 
    \caption{Comparisons between weight and transferred activation quantization errors. Statistics are extracted from the 20th channel of \texttt{blocks.3.conv2} in ResNet-18 on CIFAR-100. (a) Weight quantization error from 2-bit channel-wise weight quantization. (b) Transferred activation quantization error by Eq.~\eqref{eqt:erq} under 2-bit activation quantization.}
    \label{fig:error}
\end{figure}

We found that the transfer strategy in ERQ~\cite{10.5555/3692070.3694620} is not well suited to ETBQ. Specifically, given $\hat{\boldsymbol{a}}_l=\boldsymbol{a}_l + \delta\boldsymbol{a}_l$, the equivalent weight error $\delta\boldsymbol{W}^{\ast}$ is obtained by matching the FP output $\boldsymbol{W}\boldsymbol{a}_l$ with $(\boldsymbol{W}+\Delta)\hat{\boldsymbol{a}}_l$:
\begin{equation}
\label{eqt:erq}
    \begin{aligned}
    \mathcal{J}(\Delta)
    &=\mathbb{E}\!\left[
    \left\|\boldsymbol{W}\boldsymbol{a}_l-(\boldsymbol{W}+\Delta)\hat{\boldsymbol{a}}_l\right\|_2^2
    \right]+\lambda\|\Delta\|_F^2 .
    \end{aligned}
\end{equation}
The resulting solution can be interpreted as a transferred activation error to the weight counterpart $\delta\boldsymbol{W}^{\ast}$. Consequently, one could then use a unified weight-side error $\delta\boldsymbol{W}_{\mathrm{total}}=\delta\boldsymbol{W}+\delta\boldsymbol{W}^{\ast}$ for FP pre-conditioning.

\begin{figure}[t!]
    \centering
    \includegraphics[width=1.0\linewidth]{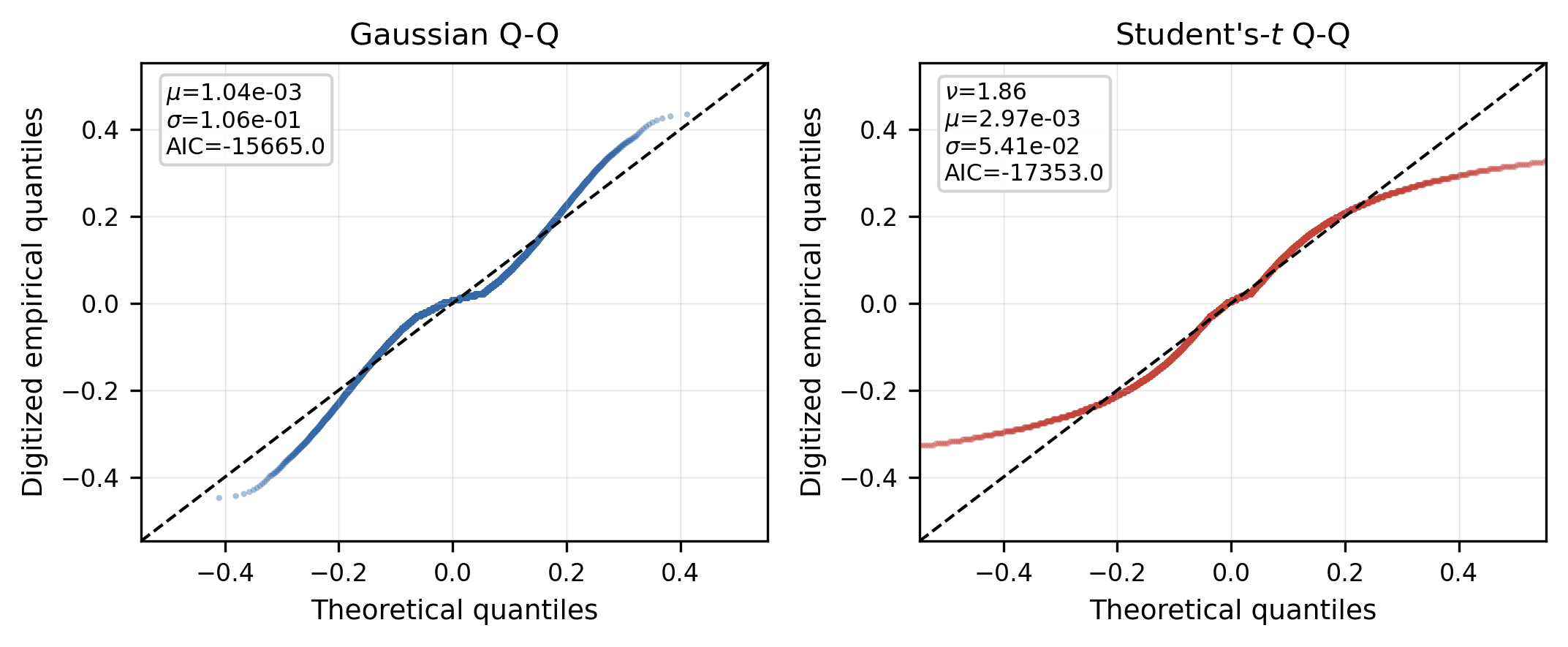}
    \caption{Q-Q diagnostics based on the digitized transferred-error histogram in Fig.~\ref{fig:error}(b). The Gaussian fit shows stronger tail deviation, whereas the Student's-$t$ fit better follows the empirical tail quantiles. This diagnostic illustrates the heavy-tailed behavior of the transferred activation error.}
    \label{fig:qq_transferred_error}
\end{figure}

However, the Frobenius regularizer in Eq.~\eqref{eqt:erq} corresponds to an isotropic zero-mean Gaussian prior on $\Delta$, encouraging small and Gaussian-like equivalent weight errors. In fact, we find that the transferred activation quantization error is not well described by a Gaussian distribution and exhibits heavy-tailed behavior. Specifically, to examine whether the two errors are directly comparable on the weight side, we empirically measured $\delta\boldsymbol{W}$ and $\delta\boldsymbol{W}^{\ast}$ on a representative convolution layer of ResNet-18 trained on CIFAR-100 Fig.~\ref{fig:error} as follows:
\begin{itemize}
    \item \textbf{Transferred activation quantization error and weight quantization one have totally different zero points.} Although the two distributions have similar near-zero means, their scales are highly different. The standard deviation of the transferred activation error is nearly two orders of magnitude larger than that of the intrinsic weight error ($1.06\times10^{-1}$ vs. $5.62\times10^{-3}$). Therefore, if $\delta\boldsymbol{W}$ and $\delta\boldsymbol{W}^{\ast}$ are simply combined, the transferred activation error dominates the total weight-side error, while the intrinsic weight quantization error becomes underrepresented.
    \item \textbf{Transferred activation quantization error overwhelms the weight quantization one.} The standard deviation of the transferred activation error is nearly two orders of magnitude larger than that of the intrinsic weight error ($1.06\!\times\!10^{-1}$ vs.\ $5.62\!\times\!10^{-3}$). If $\delta\boldsymbol{W}$ and $\delta\boldsymbol{W}^{\ast}$ are simply combined, the transferred activation error dominates the total error, while the intrinsic weight quantization error becomes underrepresented.
        \item \textbf{Transferred activation quantization error is long-tailed.} Moreover, Fig.~\ref{fig:qq_transferred_error} shows that the transferred activation error has heavy-tailed behavior and barely follows the Gaussian assumption in Eq.~\eqref{eqt:erq}.
\end{itemize}

This observation shows that the two types of error sources are not comparable in scale. ETBQ therefore adopts a joint approach: WQN conditions the FP model by channel-wise weight quantization error $\delta\boldsymbol{W}$, whereas AQN conditions the model by activation quantization error $\delta\boldsymbol{a}_l$ at activation quantizer locations. Thus, weight and activation errors are optimized together during FP pre-conditioning, but each error source is modeled and injected in its native space.

\section{Proposed Method}
\label{sec:method}

\subsection{Weight Quantization Noise}
\label{sec:wqn}

\paragraph{Quantization error modeling.}
WQN models the perturbation that the target weight quantizer will actually introduce, so that the FP model is conditioned against the genuine quantization error rather than generic perturbation. Concretely, we simulate the target PTQ weight quantizer on the current FP weights $\boldsymbol{W}_t$ and obtain the weight quantization error 
$\boldsymbol{E}_{\boldsymbol{W}_t} = \hat{\boldsymbol{W}}_t - \boldsymbol{W}_t$, where $t$ denotes the training iteration. Without loss of generality, when the weight quantizer is applied channel-wise to $\boldsymbol{W}_t$, WQN estimates the corresponding error statistics\footnote{We use error, perturbation, and noise to describe injected quantization-error samples when no ambiguity arises.} at the same granularity. In this paper, we follow previous work~\cite{weiqdrop} and the empirical error distributions reported in Appendix~\ref{app:error_distribution} to model the error as a channel-wise Gaussian distribution:
\begin{equation}
\label{eq:wqn_error}
\boldsymbol{\epsilon}_{\boldsymbol{W}_t} \sim \mathcal{N}(\boldsymbol{\mu}_{\boldsymbol{W}_t}, \boldsymbol{\Sigma}_{\boldsymbol{W}_t}),
\end{equation}
where $\boldsymbol{\mu}_{\boldsymbol{W}_t}$ and $\boldsymbol{\Sigma}_{\boldsymbol{W}_t}$ are computed from $\boldsymbol{E}_{\boldsymbol{W}_t}$. This design aligns the injected perturbation with the target weight quantizer instead of using generic isotropic perturbation. Details of how to estimate $\boldsymbol{\mu}_{\boldsymbol{W}_t}$ and $\boldsymbol{\Sigma}_{\boldsymbol{W}_t}$ are provided in Appendix~\ref{app:wqn_statistics} and Appendix~\ref{app:error_distribution}.

\paragraph{Temporal differencing for a flat solution.}
Directly adding independently sampled errors at each iteration may introduce a bias when the empirical quantization-error distribution has a non-zero mean. Repeated updates with such biased errors would cause drift in the FP optimization trajectory. Therefore, WQN uses temporal differencing. At iteration $t$, a quantization-aligned weight error is sampled as
\begin{equation}
\boldsymbol{\delta}_{\boldsymbol{W}_t} = \lambda_e \boldsymbol{\epsilon}_{\boldsymbol{W}_t},
\end{equation}
where $\lambda_e \in [0,1]$ is a warmup-controlled error intensity (see Section~\ref{sec:training_algorithm}). Instead of applying $\boldsymbol{\delta}_{\boldsymbol{W}_t}$ directly, WQN forms the differenced error
\begin{equation}
\label{eq:wqn_diff}
\boldsymbol{P}_{\boldsymbol{W}_t}
=
\boldsymbol{\delta}_{\boldsymbol{W}_t}
-
\boldsymbol{\delta}_{\boldsymbol{W}_{t-1}} .
\end{equation}
The task loss is then evaluated at the perturbed weights
\begin{equation}
\label{eq:wqn_forward}
\boldsymbol{W}'_t
=
\boldsymbol{W}_t+\boldsymbol{P}_{\boldsymbol{W}_t},
\end{equation}
and the underlying clean FP weights are updated by
\begin{equation}
\label{eq:wqn_update}
\boldsymbol{W}_{t+1}
=
\boldsymbol{W}_t
-
\eta \nabla_{\boldsymbol{W}}
\mathcal{L}_{\mathrm{task}}
\big(f(\boldsymbol{x};\boldsymbol{W}'_t),y\big).
\end{equation}
Thus, WQN changes the point at which the gradient is evaluated, but the stored model parameters remain clean FP weights.

The differenced error in Eq.~\eqref{eq:wqn_diff} has two effects. First, it suppresses the trajectory-level drift caused by non-zero-mean quantization errors, since the accumulated differenced error telescopes over iterations. Second, because the gradient is evaluated under quantization-aligned local errors, the optimizer is discouraged from solutions whose loss changes sharply under such errors. In this sense, WQN acts as an error-conditioned smoothing mechanism: sharp regions produce unstable gradients under $\boldsymbol{P}_{\boldsymbol{W}_t}$, whereas flatter regions are less affected. This encourages the FP model to move toward a flatter basin that is more tolerant to the target weight quantization error.

\paragraph{Connection to anti-correlated noise injection.}
The temporal differencing in Eq.~\eqref{eq:wqn_diff} is related to anti-correlated noise injection~\cite{orvieto2022anticorrelated}: the perturbations of consecutive iterations share a common, oppositely-signed component rather than being independent. Anti-PGD~\cite{orvieto2022anticorrelated} shows that such anti-correlated perturbations can bias gradient descent toward regions of lower loss curvature, which is equivalent to implicitly penalizing the Hessian and favoring flatter minima. We use this result as the motivation for WQN, while noting that WQN uses anisotropic and generally non-zero-mean quantization-aligned perturbations rather than the idealized isotropic perturbations in Anti-PGD. 

However, two properties distinguish WQN from generic anti-correlated injection. First, the per-step perturbation $\boldsymbol{\delta}_{\boldsymbol{W}_t}=\lambda_e\boldsymbol{\epsilon}_{\boldsymbol{W}_t}$ is \emph{quantization-aligned}: it is sampled from the channel-wise error distribution of the target weight quantizer instead of an isotropic Gaussian, so the induced smoothing concentrates along the directions in which the quantizer produces the largest error. Second, since the quantization error is not zero-mean, the differencing additionally cancels its systematic mean component and prevents it from accumulating into the FP weights, a drift-control effect that is unnecessary for the zero-mean noise used in Anti-PGD. The mechanism-level ablation in Appendix~\ref{app:wqn_temporal_differencing} empirically evaluates the correctness of the above approach.

\subsection{Activation Quantization Noise}
\label{sec:aqn}

AQN reduces the FP model’s sensitivity to activation-side quantization error $\boldsymbol{W}\delta\boldsymbol{a}_l$ (Eq.~\eqref{eqt:quantization_delta}). Activations differ from weights in two key ways: 1) they are input-dependent, so the quantization-error statistics fluctuate across mini-batches; 2) they are high-dimensional, so dense perturbations can destabilize intermediate representations during pre-conditioning.

\paragraph{Activation-error modeling.}
For each activation quantizer, we collect the FP activations $\boldsymbol{a}_l$ and their quantized counterparts $\hat{\boldsymbol{a}}_l$ on the calibration set, and define the activation quantization error as follows:
\begin{equation}
\label{eq:aqn_activation_error}
\boldsymbol{E}_{\boldsymbol{a}_l} = \hat{\boldsymbol{a}}_l - \boldsymbol{a}_l.
\end{equation}
Since activations are quantized per tensor in our PTQ setting, AQN estimates one scalar mean and variance for each activation quantizer. Specifically, for the activation quantizer at layer $l$, we model the activation quantization error by a per-tensor Gaussian distribution:
\begin{equation}
\label{eq:aqn_distribution}
\boldsymbol{E}_{\boldsymbol{a}_l} \sim \mathcal{N}(\mu_{\boldsymbol{a}_l}, \sigma_{\boldsymbol{a}_l}^{2}),
\end{equation}
where $\mu_{\boldsymbol{a}_l}\in\mathbb{R}$ and $\sigma_{\boldsymbol{a}_l}^{2}\in\mathbb{R}$ are estimated from all elements of $\boldsymbol{E}_{\boldsymbol{a}_l}$ over the calibration subset. This distribution is later used to sample the activation quantization-error tensor $\boldsymbol{\Xi}_{\boldsymbol{a}_l}$ in Eq.~\eqref{eq:aqn_forward}. The modeling is consistent with the per-tensor activation quantizer used in the downstream PTQ backend. To suppress batch-level fluctuations, $(\mu_{\boldsymbol{a}_l}, \sigma_{\boldsymbol{a}_l}^2)$ are updated across epochs using exponential moving average (Appendix~\ref{app:aqn_statistics}).

\paragraph{Salience-aware stochastic masking.}
Applying sampled activation errors to all elements may over-perturb the intermediate activations. We use the spatial magnitude of each channel as a simple salience proxy: channels with larger responses tend to have larger absolute activation quantization errors under per-tensor quantization, and perturbing them exposes the model to more influential activation-side errors. In this paper, a binary mask $\boldsymbol{M} \in \{0,1\}^{B \times C}$ is proposed to select which channels receive activation quantization-error injection. For the intermediate activation tensor $\boldsymbol{a}_l \in \mathbb{R}^{B\times C \times H \times W}$, the channel salience is computed as the spatial absolute mean:
\begin{equation}
\label{eqt:channel_salience}
\boldsymbol{I}_{b,c} = \frac{1}{H W}\sum_{h,w} |\boldsymbol{a}_{l,b,c,h,w}|.
\end{equation}
The salience scores are converted into Bernoulli probabilities with expected density controlled by hyperparameter $\rho$:
\begin{equation}
\label{eq:aqn_mask_probability}
\boldsymbol{\Pi} = \operatorname{clip}\big(C \rho \cdot \operatorname{softmax}(\boldsymbol{I}),\, 0, 1\big), \quad
\boldsymbol{M} \sim \operatorname{Bernoulli}(\boldsymbol{\Pi}),
\end{equation}
where $C$ is the number of channels and $\rho\in[0,1]$ controls the target masking density. Since $\operatorname{softmax}(\boldsymbol{I})$ sums to one across channels for each sample, the factor $C\rho$ rescales the distribution so that the expected number of selected channels is approximately $C\rho$ when clipping is inactive. The clipping operation keeps all entries of $\boldsymbol{\Pi}$ within the valid probability range $[0,1]$, and $\operatorname{Bernoulli}(\boldsymbol{\Pi})$ independently samples each channel mask according to its salience-dependent probability. The mask $\boldsymbol{M}$ is broadcast along the spatial dimensions $(H,W)$ so that all spatial locations within a selected channel share the same injection decision (Appendix~\ref{app:aqn_mask_detail}).

\paragraph{Perturbation injection.}
Inspired by~\cite{pang2025stabilizingquantizationawaretrainingimplicitregularization}, AQN injects sampled activation quantization errors rather than generic random noise. Specifically, the activation-error statistics $(\mu_{\boldsymbol{a}_l},\sigma^2_{\boldsymbol{a}_l})$ are estimated from the empirical activation quantization error $\boldsymbol{E}_{\boldsymbol{a}_l}$ on the calibration subset. During pre-conditioning, we sample an activation quantization-error tensor $\boldsymbol{\Xi}_{\boldsymbol{a}_l}\in\mathbb{R}^{B\times C\times H\times W}$, whose elements are independently drawn from the estimated distribution:
\begin{equation}
\boldsymbol{\Xi}_{\boldsymbol{a}_l}^{b,c,h,w} \sim \mathcal{N}(\mu_{\boldsymbol{a}_l},\sigma_{\boldsymbol{a}_l}^{2}).
\end{equation}
The final perturbed activation is then computed as
\begin{equation}
\label{eq:aqn_forward}
\boldsymbol{a}_l' = \boldsymbol{a}_l + \lambda_e \left(\boldsymbol{M}^{\uparrow} \odot \boldsymbol{\Xi}_{\boldsymbol{a}_l} \right),
\end{equation}
where $\lambda_e$ controls the activation-error injection strength, and $\boldsymbol{M}^{\uparrow}\in\{0,1\}^{B\times C\times H\times W}$ is the spatially broadcast version of $\boldsymbol{M}\in\{0,1\}^{B\times C}$.

\begin{figure}[t!] 
\centering \includegraphics[width=1\linewidth]{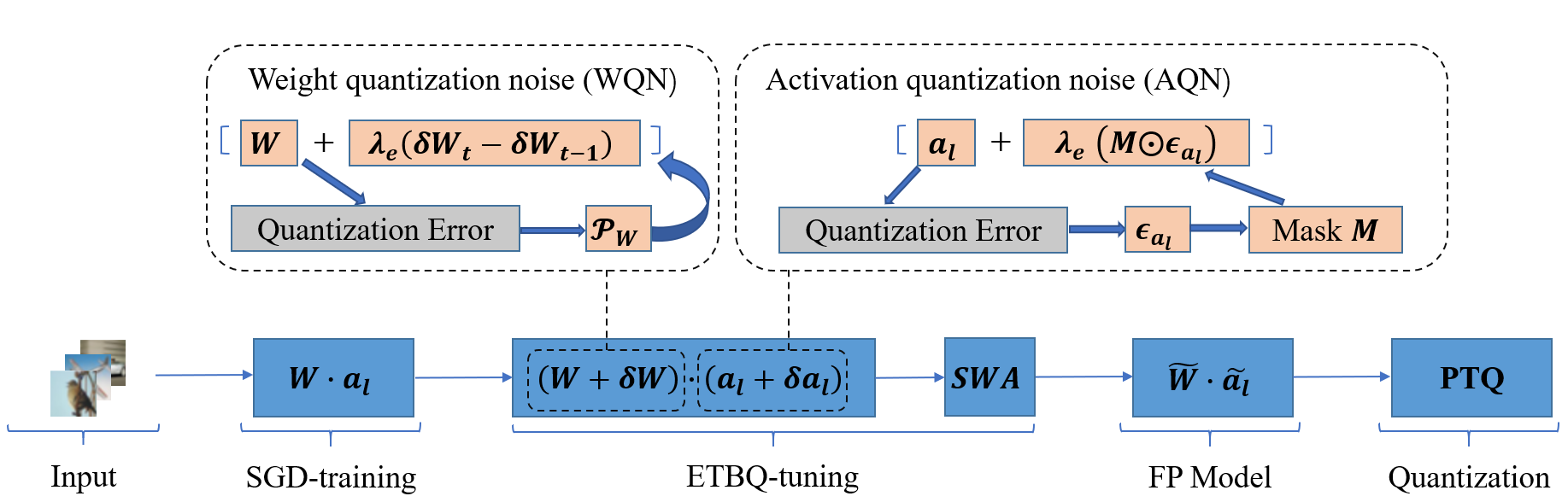} \caption{ETBQ inserts a pre-conditioning stage between FP training and PTQ. The stage injects both weight-side and activation-side errors conditioning through WQN and AQN, respectively.}
\label{fig:framework} 
\end{figure}

The overall ETBQ framework, integrating WQN and AQN into a unified pre-conditioning pipeline, is summarized in Fig.~\ref{fig:framework}.

\subsection{Training Objective and Algorithm}
\label{sec:training_algorithm}

For a minibatch $(\boldsymbol{x}_0,y) \in \mathscr{D}_t$, let $\mathcal{L}_\mathrm{task}$ denote the task-specific loss. During ETBQ tuning, the network forward pass applies:
\begin{itemize}
    \item WQN-perturbed weights $\boldsymbol{W}'_t = \boldsymbol{W}_t + \boldsymbol{P}_{\boldsymbol{W}_t}$ (Eq.~\eqref{eq:wqn_forward}); and
    \item AQN-perturbed intermediate activations $\boldsymbol{a}_l' = \boldsymbol{a}_l + \lambda_e(\boldsymbol{M}^{\uparrow}\odot\boldsymbol{\Xi}_{\boldsymbol{a}_l})$ (Eq.~\eqref{eq:aqn_forward}), applied at all activation quantizer locations.
\end{itemize}

The training objective is
\begin{equation}
\label{eq:etbq_loss}
\mathcal{L}_{\mathrm{ETBQ}} = \mathcal{L}_\mathrm{task}\big(\tilde{f}(\boldsymbol{x}_0; \boldsymbol{W}'_t), y\big),
\end{equation}
where $\tilde{f}$ denotes the forward pass under AQN-perturbed activations.

The quantization error intensity $\lambda_e$ follows a linear warmup schedule:
\begin{equation}
\label{eqt:warmup_schedule}
\lambda_e = \min\!\Big(\frac{e}{E_\mathrm{warmup}}, 1\Big) \cdot \lambda_\mathrm{max},
\end{equation}
where $E_\mathrm{warmup}$ is the warmup length in epochs and $\lambda_\mathrm{max}$ is the target quantization error intensity. The warmup ensures that the FP model is gradually exposed to the target perturbations without destabilizing the early training stage.

At the final stage of ETBQ, stochastic weight averaging (SWA) consolidates the late-stage trajectory to produce a more stable FP model. Since SWA changes the effective parameters, BatchNorm statistics are recomputed before exporting the model for PTQ. Algorithm~\ref{alg:etbq_corrected} summarizes the ETBQ pre-conditioning procedure.

\begin{algorithm}[t!]
   \caption{The ETBQ Framework}
   \label{alg:etbq_corrected}
\begin{algorithmic}[1]
   \STATE {\bfseries Input:} Pre-trained FP model $f(\cdot;\boldsymbol{W})$, training set $\mathscr{D}_t$, calibration set $\mathscr{D}_c$, total epochs $E$, SWA start epoch $E_{\mathrm{swa}}$.
   \STATE {\bfseries Initialize:} 
        Estimate initial WQN and AQN statistics on $\mathscr{D}_c$; 
        set previous raw weight perturbation $\boldsymbol{\delta}_{w,0}\leftarrow \mathbf{0}$.
   \FOR{$e=1$ {\bfseries to} $E$}
        \STATE Update the quantization error intensity $\lambda_e$ using Eq.~\eqref{eqt:warmup_schedule}.
        \STATE Refresh WQN and AQN statistics on $\mathscr{D}_c$ (Appendices~\ref{app:wqn_statistics} and~\ref{app:aqn_statistics}).

        \FOR{each minibatch $(\boldsymbol{x}_0,y) \in \mathscr{D}_t$}
            \STATE Sample the raw WQN perturbation \(\boldsymbol{\delta}_{\boldsymbol{W}_t}\).
            \STATE Compute the WQN differenced perturbation $\boldsymbol{P}_{\boldsymbol{W}_t}$ (Eq.~\eqref{eq:wqn_diff}).
            \STATE Forward pass using WQN-perturbed weights $\boldsymbol{W}'_t$ (Eq.~\eqref{eq:wqn_forward}).
            \STATE Apply AQN-perturbed activations $\boldsymbol{a}_l'$ at all activation quantizer locations (Eq.~\eqref{eq:aqn_forward}).
            \STATE Compute ETBQ loss $\mathcal{L}_\mathrm{ETBQ}$ (Eq.~\eqref{eq:etbq_loss}) and update FP weights $\boldsymbol{W}_t$ via SGD.
            \STATE Store the current raw perturbation $\boldsymbol{\delta}_{\boldsymbol{W}_t}$ for the next differential WQN step.
        \ENDFOR

        \IF{$e \geq E_{\mathrm{swa}}$}
            \STATE Update SWA weights: $\boldsymbol{W}_\mathrm{swa} \leftarrow \mathrm{SWA}(\boldsymbol{W}_t)$.
        \ENDIF
   \ENDFOR

   \STATE Recompute BatchNorm statistics for $\boldsymbol{W}_\mathrm{swa}$.
   \STATE {\bfseries Output:} Pre-conditioned FP model $\boldsymbol{W}_\mathrm{swa}$, ready for any reconstruction-based PTQ backend.
\end{algorithmic}
\end{algorithm}

\section{Experiments}
\subsection{Implementation Details}

\textbf{Models and Datasets.}
We evaluate ETBQ on image classification and semantic segmentation. For classification, we use CIFAR-100~\cite{krizhevsky-cifar-arxiv-2009}, Tiny-ImageNet~\cite{le2015tiny}, and ImageNet~\cite{russakovsky2015imagenet}. On CIFAR-100, we evaluate ResNet-18 and ResNet-50~\cite{he-resnet-cvpr-2016} as standard convolutional networks, and MobileNet-V1~\cite{howard-mbv1-arxiv-2017} and MobileNet-V2~\cite{sandler-mbv2-cvpr-2018} as lightweight architectures. On Tiny-ImageNet, we evaluate ResNet-18 and GhostNet~\cite{han2020ghostnet} to test generality across model families. On ImageNet, we evaluate ResNet-18, ResNet-50, MobileNet-V1, and MobileNet-V2 to verify scalability on large-scale classification. For semantic segmentation, we evaluate U-Net~\cite{ronneberger2015u} on Cityscapes~\cite{cordts2016cityscapes}, where dense spatial prediction makes activation quantization particularly challenging. For PTQ calibration, we randomly sample 100 images from CIFAR-100, 200 images from Tiny-ImageNet, 1024 images from ImageNet, and 128 images from Cityscapes; each calibration set is kept fixed across all compared PTQ methods on the corresponding dataset.

\textbf{Pre-conditioning Cost.}
ETBQ introduces an additional FP pre-conditioning stage before PTQ. On CIFAR-100, the model is tuned for 120 epochs with batch size 256; the quantization error intensity warms up for 20 epochs and SWA is applied during the final 60 epochs. On ImageNet, ETBQ is tuned for 80 epochs as reported in Appendix~\ref{app:additional_ablation}. The same training data are used for FP pre-conditioning, while WQN/AQN statistics and PTQ calibration are estimated from the fixed calibration subset. 
The purpose of this protocol is to isolate the effect of improving the FP starting point for PTQ, not to claim that ETBQ dominates all possible equal-budget QAT fine-tuning variants.

\textbf{Experimental Setup.}
All experiments are implemented in PyTorch~\cite{paszke-pytorch-nips-2019} with MQBench~\cite{li-mqbench-arxiv-2021}. Unless otherwise specified, we use QDrop~\cite{weiqdrop} as the reconstruction-based PTQ procedure with asymmetric uniform quantization. The first and last layers are kept at 8-bit precision to avoid excessive information loss at network boundaries. We use per-channel quantization for weights and per-tensor quantization for activations. Accordingly, WQN estimates channel-wise weight-error statistics, while AQN estimates per-tensor activation-error statistics. The salience-aware channel mask in AQN is used only to select perturbation locations during FP pre-conditioning and does not change the downstream activation quantizer granularity. We set the maximum quantization error intensity to $\lambda_{\max}=1.0$ unless otherwise specified, update AQN statistics with EMA momentum $\beta=0.9$, and recompute BatchNorm statistics after SWA before PTQ calibration. We use W$X$A$Y$ to denote $X$-bit weight and $Y$-bit activation quantization. For ETBQ pre-conditioning, the FP model is tuned before PTQ using the task-specific training loss, while WQN/AQN statistics are estimated from the calibration subset. We report both absolute quantized performance and the quantization drop relative to the corresponding FP32 model, because ETBQ may also change the FP32 starting point. For classification, we report top-1 accuracy; for segmentation, we report mean Intersection-over-Union (mIoU). All ETBQ experiments are conducted on a single NVIDIA RTX 2080 Ti GPU with an Intel i9-9900KF CPU.

\textbf{Compared Methods.}
We compare ETBQ with representative reconstruction-based PTQ methods, including AdaRound~\cite{nagel-adaround-icml-2020}, BRECQ~\cite{librecq}, QDrop~\cite{weiqdrop}, and QEP~\cite{arai2026quantizationerrorpropagationrevisiting}. All baselines are applied to the standard SGD-trained FP model using the same calibration samples, bit-widths, and quantization settings whenever applicable. Since ETBQ is an upstream pre-conditioning stage rather than a PTQ backend, we use QDrop as the default downstream PTQ method and additionally report QEP compatibility. Thus, the direct PTQ baseline denotes QDrop applied to the original FP model, whereas ETBQ denotes the same QDrop procedure applied to the ETBQ-preconditioned FP model. This protocol isolates the effect of tuning before quantization from changes in the PTQ algorithm.

\subsection{Ablation Study}
\label{sec:ablation}

We conduct ablation studies on CIFAR-100 with ResNet-18~\cite{he-resnet-cvpr-2016} under the W2A4 setting. The baseline is the SGD-trained FP model followed by QDrop. For ETBQ tuning, we use SGD with momentum 0.9, batch size 256, weight decay 0.0005, and an initial learning rate of 0.015 with cosine annealing. The model is tuned for 120 epochs. The quantization error intensity $\lambda_e$ linearly increases from 0 to $\lambda_{\max}=1.0$ over the first 20 epochs and is then kept fixed. SWA is applied during the final 60 epochs. We use the task-specific loss with label smoothing factor 0.1. The main text reports component-level ablations, while mechanism-level ablations of WQN temporal differencing and AQN salience-aware masking are provided in Appendix~\ref{app:additional_ablation}.

\begin{table}[h]
\centering
\caption{Ablation of ETBQ on ResNet-18 under W2A4 on CIFAR-100 (\%).}
\label{tab:ablation_components}
\resizebox{0.45\textwidth}{!}{
\begin{tabular}{ccc|ccc}
\toprule
\textbf{WQN} & \textbf{AQN} & \textbf{SWA} & \textbf{FP32} & \textbf{W2A4 } & \textbf{Quant. Drop} \\
\midrule
-- & -- & -- & 79.40 & 76.47 & -2.93 \\
\checkmark & -- & -- & 79.73 & 77.67 & -2.06 \\
-- & \checkmark & -- & 79.98 & 77.37 & -2.61 \\
-- & -- & \checkmark & 78.79 & 77.22 & -1.57 \\
-- & \checkmark & \checkmark & 79.76 & 77.27 & -2.49 \\
\checkmark & \checkmark & -- & 79.56 & 77.43 & -2.13 \\
\checkmark & -- & \checkmark & \textbf{79.98} & 77.74 & -2.24 \\
\checkmark & \checkmark & \checkmark & 79.81 & \textbf{78.06} & -1.75 \\
\bottomrule
\end{tabular}
}
\end{table}

Table~\ref{tab:ablation_components} shows that each component contributes to low-bit quantization robustness. The direct QDrop baseline drops from 79.40\% FP32 accuracy to 76.47\% under W2A4 quantization. Adding WQN alone improves W2A4 accuracy to 77.67\%, giving the largest single-component gain. This supports the role of weight-side quantization-error conditioning: WQN optimizes the FP model under errors aligned with the estimated weight quantization-error distribution rather than under generic random perturbations.

AQN alone improves W2A4 accuracy from 76.47\% to 77.37\%, suggesting that activation-side perturbation conditioning is also beneficial. SWA alone reaches 77.22\%, indicating that trajectory averaging improves quantization tolerance even without explicit quantization-error modeling. This means that a flat solution from SWA is beneficial for quantization. We have two interesting observations as follows:
\begin{itemize}
    \item \textbf{Flatness is critical for quantization.} Interestingly, the WQN+AQN without SWA (77.43\%) does not outperform WQN (77.67\%) alone, suggesting that simultaneous weight and activation perturbations produces non-smooth SGD trajectory. When SWA is added, the full ETBQ achieves the best W2A4 accuracy of 78.06\%, showing that SWA helps consolidate the perturbation-driven trajectory into a more stable FP solution.
    \item \textbf{High FP32 accuracy does not guarantee strong performance after quantization.} Specifically, WQN+SWA achieves an FP32 accuracy of 79.98\%, outperforming WQA+AQN+SWA (ETBQ), which records 79.81\% at FP32 precision. Upon quantization, however, ETBQ surpasses its WQN+SWA counterpart by 0.28\%.
\end{itemize}
This suggests that SWA helps consolidate the error-conditioned tuning trajectory into a flatter and more stable solution.

The performance gains of ETBQ may come from two sources: improved FP32 accuracy and reduced quantization drop. The full ETBQ improves W2A4 accuracy by 1.59\% over the direct PTQ baseline, of which 0.41\% comes from FP32 improvement and 1.18\% comes from reducing the quantization drop. This indicates that ETBQ does not merely improve FP accuracy; its main contribution is improving robustness to subsequent low-bit quantization.

\subsection{Performance gains to PTQ Methods}
\label{sec:comparison}

We applied ETBQ with representative PTQ methods on image classification and semantic segmentation. Since ETBQ is an upstream tuning-before-quantization stage, it is not intended to replace existing PTQ backends. Unless otherwise specified, QDrop and QEP are used as the downstream representative PTQ methods after ETBQ pre-conditioning. Therefore, the comparison between direct QDrop(QEP) and ETBQ isolates the effect of improving the FP model before quantization.

\subsubsection{Image Classification on CIFAR-100}

Table~\ref{tab:sota_comparison_cifar100} reports CIFAR-100 results across ResNet-18, ResNet-50, MobileNet-V1, and MobileNet-V2. ETBQ achieves the highest absolute quantized accuracy across all architectures and bit-widths. Compared with direct QDrop, ETBQ improves W4A4 accuracy by 1.07\%, 0.52\%, 1.02\%, and 1.53\% on ResNet-18, ResNet-50, MobileNet-V1, and MobileNet-V2, respectively. These results indicate that ETBQ benefits both standard residual networks and lightweight architectures.

The advantage of ETBQ is significant when low bit quantization is used. For example, ETBQ improves ResNet-50 over direct QDrop by 2.34\% under W2A4 and 3.03\% under W2A2. On MobileNet-V2, ETBQ improves W2A2 accuracy from 20.63\% to 23.89\%. These results verify that pre-conditioning the FP model is particularly useful for the low-bit quantization setting. It means that ETBQ is an efficient method to obtain a pre-conditioned FP model. The only case with a larger relative drop is ResNet-50 W4A4; however, ETBQ still achieves the highest absolute accuracy.

\begin{table}[t!]
\centering
\caption{Comparison between PTQ and ETBQ-preconditioned PTQ on CIFAR-100. Results are top-1 accuracy (\%); values in parentheses denote quantization drop relative to each method's own FP32 model. For each architecture, all PTQ baselines share the same SGD-pretrained FP32 model shown in the shared FP32 cell, while ETBQ rows use the corresponding ETBQ-preconditioned FP32 model.}
\label{tab:sota_comparison_cifar100}
\resizebox{0.45\textwidth}{!}{
\begin{tabular}{lccccc}
\toprule
\multirow{2}{*}{\textbf{Architecture}} & \multirow{2}{*}{\textbf{Method}} & \multirow{2}{*}{\textbf{FP32 Acc. (\%)}} & \multicolumn{3}{c}{\textbf{Quantized Accuracy (\%)}} \\

& & & \textbf{W4A4} & \textbf{W2A4} & \textbf{W2A2} \\
\midrule
\multirow{6}{*}{ResNet-18}
& AdaRound~\cite{nagel-adaround-icml-2020} & \multirow{4}{*}{79.40} & 77.46 (-1.94) & 75.25 (-4.15) & 70.70 (-8.70) \\
& BRECQ~\cite{librecq}         &                        & 77.61 (-1.79) & 76.52 (-2.88) & 72.13 (-7.27) \\
& QDrop~\cite{weiqdrop}        &                        & 77.79 (-1.61) & 76.41 (-2.99) & 72.29 (-7.11) \\
& QEP~\cite{arai2026quantizationerrorpropagationrevisiting} &  & 78.21 (-1.19) & 77.38 (-2.02) & 71.91 (-7.49)\\
& ETBQ (Ours) + QDrop~\cite{weiqdrop}               & \multirow{2}{*}{\textbf{79.81}}         & \textbf{78.86 (-0.95)} & \textbf{78.06 (-1.75)} & \textbf{73.95 (-5.86)} \\
& ETBQ (Ours) + QEP~\cite{arai2026quantizationerrorpropagationrevisiting}  & 
& 78.68 (-1.13) & 77.51 (-2.30) & 73.04 (-6.77)\\
\midrule
\multirow{6}{*}{ResNet-50}
& AdaRound~\cite{nagel-adaround-icml-2020} & \multirow{4}{*}{78.94} & 77.13 (-1.81) & 75.78 (-3.16) & 68.95 (-9.99) \\
& BRECQ~\cite{librecq}         &                        & 78.52 (-0.42) & 76.32 (-2.62) & 70.61 (-8.33) \\
& QDrop~\cite{weiqdrop}        &                        & 78.71 (-0.23) & 76.75 (-2.19) & 71.73 (-7.21) \\
& QEP~\cite{arai2026quantizationerrorpropagationrevisiting} &  & 78.86 (-0.08) & 76.63 (-2.31) & 71.89 (-7.05)  \\
& ETBQ (Ours) + QDrop~\cite{weiqdrop}              & \multirow{2}{*}{\textbf{80.82}}         & 79.23 (-1.59) & 79.09 (-1.73) & 74.76 (-6.06) \\
& ETBQ (Ours) + QEP~\cite{arai2026quantizationerrorpropagationrevisiting}  & 
& \textbf{81.90 (+1.08)}$^\dagger$ & \textbf{80.79 (-0.03)} & \textbf{75.25 (-5.57)} \\
\midrule
\multirow{6}{*}{MobileNet-V1}
& AdaRound~\cite{nagel-adaround-icml-2020} & \multirow{4}{*}{70.22} & 64.40 (-5.82) & 33.35 (-36.87) & 12.60 (-57.62) \\
& BRECQ~\cite{librecq}         &                        & 65.92 (-4.30) & 54.03 (-16.19) & 20.09 (-50.13) \\
& QDrop~\cite{weiqdrop}        &                        & 66.81 (-3.41) & 62.57 \textbf{(-7.65)} & 46.40 (-23.82) \\
& QEP~\cite{arai2026quantizationerrorpropagationrevisiting} &  & 66.83 (-3.39) & 62.48 (-7.74) & 46.12 (-24.09) \\
& ETBQ (Ours) + QDrop~\cite{weiqdrop}                     & \multirow{2}{*}{\textbf{70.94}}         & \textbf{67.83 (-3.11)} & 63.28 (-7.66) & \textbf{47.23 (-23.71)} \\
& ETBQ (Ours) + QEP~\cite{arai2026quantizationerrorpropagationrevisiting}  & 
& 67.81 (-3.13) & \textbf{63.47 (-7.47)} & 47.19 (-23.75) \\
\midrule
\multirow{6}{*}{MobileNet-V2}
& AdaRound~\cite{nagel-adaround-icml-2020} & \multirow{4}{*}{76.26} & 63.32 (-12.94) & 41.64 (-34.62) &  1.95 (-74.31) \\
& BRECQ~\cite{librecq}         &                        & 67.24 (-9.02) & 56.21 (-20.05) &  7.03 (-69.23) \\
& QDrop~\cite{weiqdrop}        &                        & 70.59 (-5.67) & 68.13 (-8.13) & 20.63 (-55.63) \\
& QEP~\cite{arai2026quantizationerrorpropagationrevisiting} &  & 70.88 (-5.38) & 62.44 (-13.82) & 19.23 (-57.03) \\
& ETBQ (Ours) + QDrop~\cite{weiqdrop}            & \multirow{2}{*}{\textbf{76.52}}         & 72.12 (-4.40) & 69.03 (-7.49) & \textbf{23.89 (-52.63)} \\
& ETBQ (Ours) + QEP~\cite{arai2026quantizationerrorpropagationrevisiting}  & 
& \textbf{72.47 (-4.05)} & \textbf{69.41 (-7.11)} & 23.13 (-53.39) \\
\bottomrule
\end{tabular}   
}
\vspace{1mm}
\footnotesize{$^\dagger$This W4A4 result slightly exceeds the corresponding FP32 accuracy after QEP reconstruction; we report it as an exceptional calibration outcome and use each method's own FP32 model when computing the drop.}
\end{table}

\subsubsection{Generalization on Tiny-ImageNet}
Table~\ref{tbl:tiny-imagenet} evaluates ETBQ on Tiny-ImageNet. On ResNet-18, ETBQ improves W4A4, W2A4, and W2A2 accuracies by 0.65\%, 2.14\%, and 0.84\%, respectively. The largest gain appears under W2A4, where quantization perturbations are severe but the model has not collapsed. ETBQ also improves GhostNet, a lightweight architecture that is more fragile under low-bit quantization. Under W4A4, accuracy increases from 59.01\% to 62.51\%. Under W2A2, both direct PTQ and ETBQ degrade substantially, with ETBQ improving accuracy only from 13.12\% to 14.28\%. This suggests a practical robustness boundary for extremely lightweight architectures under 2-bit weight and activation quantization.

\begin{table}[t!]
\centering
\caption{Top-1 accuracy (\%) on Tiny-ImageNet. Drops are computed relative to each method's own FP32 model.}
\label{tbl:tiny-imagenet}
\resizebox{0.95\linewidth}{!}{%
\begin{tabular}{lccccc}
\toprule
\textbf{Architecture} & \textbf{W/A} &  \textbf{SGD } & \textbf{Drop} & \textbf{ETBQ} & \textbf{Drop}\\
\midrule
\multirow{4}{*}{ResNet-18}& 32/32 & 67.92 & --- & 67.97 & --- \\
 & 4/4 & 66.85 & -1.07  & \textbf{67.50} & -0.47 \\
 & 2/4 & 63.71 & -4.21  & \textbf{65.85} & -2.12 \\
 & 2/2 & 60.51 & -7.41  & \textbf{61.35} & -6.62 \\
\midrule
\multirow{4}{*}{GhostNet (1.0x)}& 32/32 & 77.59 & --- & \textbf{77.72} & --- \\
 & 4/4 & 59.01 & -18.58  & \textbf{62.51} & -15.21\\
 & 2/4 & 35.96 & -41.63  & \textbf{37.02} & -40.70\\
 & 2/2 & 13.12  & -64.47  & \textbf{14.28} & -63.44\\
\bottomrule
\end{tabular}%
}
\end{table}

\subsubsection{Image Classification on ImageNet}

Table~\ref{tab:sota_comparison_imagenet} reports ImageNet results across ResNet-18, ResNet-50, MobileNet-V1, and MobileNet-V2.

\begin{table}[h]
\centering
\caption{Comparison between direct PTQ and ETBQ-preconditioned PTQ on ImageNet. Results are top-1 accuracy (\%); values in parentheses denote quantization drop relative to each method's own FP32 model. For each architecture, QDrop and QEP share the same SGD-pretrained FP32 model shown in the shared FP32 cell, while ETBQ rows use the corresponding ETBQ-preconditioned FP32 model.}
\label{tab:sota_comparison_imagenet}
\resizebox{0.45\textwidth}{!}{
\begin{tabular}{lccccc}
\toprule
\multirow{2}{*}{\textbf{Architecture}} & \multirow{2}{*}{\textbf{Method}} & \multirow{2}{*}{\textbf{FP32 Acc. (\%)}} & \multicolumn{3}{c}{\textbf{Quantized Accuracy (\%)}} \\

& & & \textbf{W4A4} & \textbf{W2A4} & \textbf{W2A2} \\
\midrule
\multirow{4}{*}{ResNet-18}
& QDrop~\cite{weiqdrop}        &  \multirow{2}{*}{\textbf{69.77}}  & 67.83 (-1.94) & 63.86 (-5.91) &  51.62 (-18.15) \\
& QEP~\cite{arai2026quantizationerrorpropagationrevisiting}  &   & 67.68 (-2.09) & 63.71 (-6.06) & 51.12 (-18.65) \\
& ETBQ (Ours) + QDrop~\cite{weiqdrop}               & \multirow{2}{*}{69.64}     & \textbf{68.46} (-1.18) & \textbf{64.62} (-5.02) & 52.06 (-17.58)  \\
& ETBQ (Ours) + QEP~\cite{arai2026quantizationerrorpropagationrevisiting}  &  & 68.21 (-1.43)  & 64.61 (-5.03) & \textbf{52.17} (-17.47) \\
\midrule
\multirow{4}{*}{ResNet-50}
& QDrop~\cite{weiqdrop}        &   \multirow{2}{*}{76.13}   & 74.29 (-1.84) & 70.73 (-5.40) & 59.06 (-17.07) \\
& QEP~\cite{arai2026quantizationerrorpropagationrevisiting} &   & 74.41 (-1.72) & 70.86 (-5.27) & 58.79 (-17.34) \\
& ETBQ (Ours) + QDrop~\cite{weiqdrop}               & \multirow{2}{*}{\textbf{76.45}}     & 75.35 (-1.10) & 71.88 (-4.57) & 60.23 (-16.22)  \\
& ETBQ (Ours) + QEP~\cite{arai2026quantizationerrorpropagationrevisiting}  &   & \textbf{75.97} (-0.48) & \textbf{72.13} (-4.32) & \textbf{60.45} (-16.00) \\
\midrule
\multirow{4}{*}{MobileNet-V1}
& QDrop~\cite{weiqdrop}        &  \multirow{2}{*}{70.61}   & 60.71 (-9.90) & 12.23 (-58.38) & 0.21 (-70.40) \\
& QEP~\cite{arai2026quantizationerrorpropagationrevisiting} &   & 61.58 (-9.03) & 14.21 (-56.40) & 0.65 (-69.96) \\
& ETBQ (Ours) + QDrop~\cite{weiqdrop}               & \multirow{2}{*}{\textbf{70.97}}     & \textbf{64.26} (-6.71) & \textbf{18.43} (-52.54) & 3.05 (-67.92)  \\
& ETBQ (Ours) + QEP~\cite{arai2026quantizationerrorpropagationrevisiting}  &   & 63.66 (-7.31) & 17.65 (-53.32) & \textbf{3.71} (-67.26) \\
\midrule
\multirow{4}{*}{MobileNet-V2}
& QDrop~\cite{weiqdrop}        & \multirow{2}{*}{\textbf{71.88}}     & 58.57 (-13.31) & 41.11 (-30.77) &  6.12 (-65.76) \\
& QEP~\cite{arai2026quantizationerrorpropagationrevisiting} &   & 59.33 (-12.55) & 42.09 (-29.79) & 6.54 (-65.34) \\
& ETBQ (Ours) + QDrop~\cite{weiqdrop}               & \multirow{2}{*}{71.60}     & 64.81 (-6.79) & 49.64 (-21.96) &  \textbf{8.91} (-62.69) \\
& ETBQ (Ours) + QEP~\cite{arai2026quantizationerrorpropagationrevisiting}  &   & \textbf{65.48} (-6.12) & \textbf{50.96} (-20.64) & 8.52 (-63.08) \\
\bottomrule
\end{tabular}
}
\end{table}

Table~\ref{tab:sota_comparison_imagenet} shows that ETBQ consistently improves quantized ImageNet accuracy under both QDrop and QEP backends. On standard residual networks, ETBQ+QDrop improves ResNet-18 by 0.63\%, 0.76\%, and 0.44\% under W4A4, W2A4, and W2A2, respectively; for ResNet-50, the corresponding gains are 1.06\%, 1.15\%, and 1.17\%. These gains are accompanied by smaller quantization drops, indicating that the improvement is not merely due to a different FP32 starting point but also to improved robustness after quantization.

The benefit is more pronounced on lightweight architectures, where activation and depthwise-convolution quantization are typically more fragile. On MobileNet-V1, ETBQ+QDrop improves W4A4 and W2A4 accuracy by 3.55\% and 6.20\%, respectively. On MobileNet-V2, the gains are larger, reaching 6.24\% under W4A4 and 8.53\% under W2A4. ETBQ also improves the QEP backend on both MobileNet architectures, supporting its compatibility with different reconstruction-based PTQ procedures. Nevertheless, the extremely low W2A2 setting remains challenging, especially for MobileNet-V1/V2, where the absolute accuracy is still low despite improvement. This suggests that ETBQ improves the FP model's tolerance to low-bit perturbations, but cannot fully recover information lost under extremely aggressive activation and weight quantization.

\subsubsection{Semantic Segmentation on Cityscapes}

Table~\ref{tbl:segment} evaluates U-Net on Cityscapes. ETBQ improves the FP32 mIoU from 68.52\% to 69.86\%, suggesting that pre-conditioning also acts as a regularizer for dense prediction. Under W4A4, ETBQ reaches 70.03\% mIoU, outperforming direct PTQ by 3.58\%. The gain is larger under W2A4, where ETBQ improves mIoU from 57.72\% to 63.52\%. This result suggests that ETBQ is useful for dense prediction, where activation perturbations can strongly affect spatial feature consistency. Under W2A2, the improvement becomes marginal, indicating that extremely aggressive quantization can exceed the robustness range learned during ETBQ tuning.

\begin{table}[t!]
\centering
\caption{Semantic segmentation results on Cityscapes. The metric is mIoU (\%).}
\label{tbl:segment}
\resizebox{0.95\linewidth}{!}{%
\begin{tabular}{lccccc}
\toprule
\textbf{Architecture} & \textbf{W/A} &  \textbf{SGD (baseline)} & \textbf{Drop} & \textbf{ETBQ (ours)} & \textbf{Drop} \\
\midrule
\multirow{4}{*}{U-Net} & 32/32 & 68.52 & --- & \textbf{69.86} & --- \\
 & 4/4 & 66.45 & -2.07 & \textbf{70.03} & 0.17  \\
 & 2/4 & 57.72 & -10.80 & \textbf{63.52} &  -6.34 \\
 & 2/2 & 34.29 & -34.23 & \textbf{34.88} & -34.98  \\
\bottomrule
\end{tabular}%
}
\end{table}

\begin{figure}[h]
    \centering
    \includegraphics[width=0.8\linewidth]{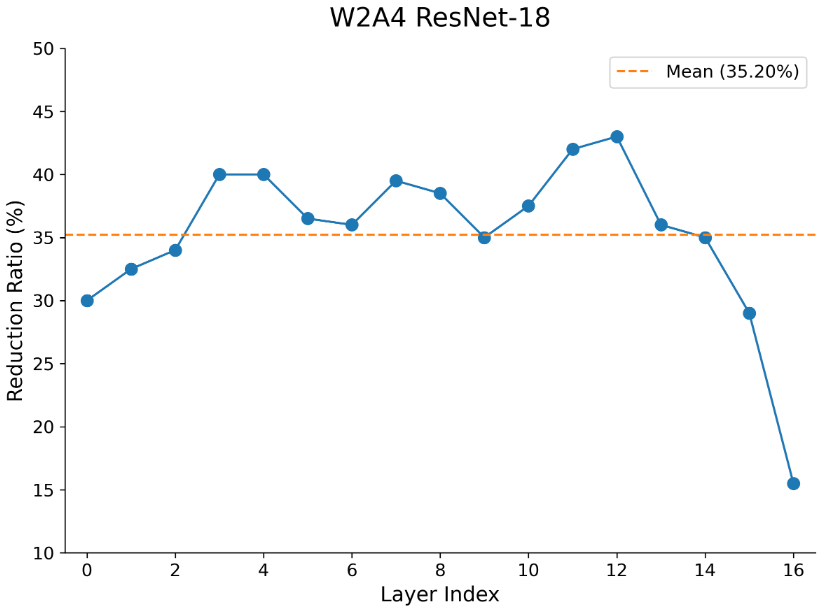}
    \caption{Layer-wise quantization error reduction ratio of ETBQ on ResNet-18 under W2A4 quantization. The dashed line denotes the average reduction ratio across layers.}
    \label{fig:resnet18-reduce}
\end{figure}

\subsection{Analysis of Quantization Error Reduction}
\label{sec:error_reduction}

To further analyze how ETBQ empowers the performance of quantization models, we measure the layer-wise reduction ratio of quantization-induced reconstruction error on ResNet-18 under W2A4 quantization.  For layer $l$, let $\boldsymbol{y}_l^{\mathrm{FP}}$ denote the full-precision layer output and $\boldsymbol{y}_l^{\mathrm{Q}}$ denote the corresponding quantized output under the same input. We define the layer-wise reconstruction error as the mean squared error (MSE) between these two outputs:
\begin{equation}
\label{eq:layer_reconstruction_error}
\mathcal{E}_l =
\mathbb{E}_{\boldsymbol{x}\in\mathscr{D}_c}
\left[
\left\|
\boldsymbol{y}_l^{\mathrm{FP}}(\boldsymbol{x})
-
\boldsymbol{y}_l^{\mathrm{Q}}(\boldsymbol{x})
\right\|_2^2
\right],
\end{equation}
where the expectation is computed over the calibration subset $\mathscr{D}_c$. We compute this error for both the original SGD-trained FP model and the ETBQ-preconditioned FP model after applying the same W2A4 PTQ backend. The reduction ratio at layer $l$ is then defined as
\begin{equation}
\label{eq:layer_error_reduction_ratio}
R_l =
\frac{
\mathcal{E}_l^{\mathrm{SGD}}
-
\mathcal{E}_l^{\mathrm{ETBQ}}
}{
\mathcal{E}_l^{\mathrm{SGD}}
}
\times 100\% .
\end{equation}

As shown in Fig.~\ref{fig:resnet18-reduce}, ETBQ reduces the output reconstruction error across most layers, with an average reduction ratio of 35.20\%. This indicates that the improvement is distributed throughout the network rather than concentrated in a single layer. The reduction is particularly pronounced in several intermediate and deeper layers, where low-bit weight and activation errors can accumulate along the forward propagation path. In contrast, the reduction becomes smaller in the final layers, suggesting that highly compressed late-stage representations remain more difficult to stabilize under aggressive quantization. Overall, these results support the view that ETBQ improves the quantization-friendliness of the FP model itself, rather than only improving calibration or reconstruction at isolated layers.

\subsection{Computation Cost}
\label{sec:limitations}

\begin{table}[t!]
\centering
\caption{Computational cost comparison on ResNet-18/ImageNet. Runtime is measured on a single NVIDIA RTX 2080 Ti. ETBQ reports the pre-conditioning cost plus the downstream QDrop calibration cost.}
\label{tbl:cost_comparison}
\resizebox{0.95\linewidth}{!}{
\begin{tabular}{lccccc}
\toprule
\textbf{Method} & \textbf{Train FP Weights} & \textbf{Optimize Quant. Params} & \textbf{Epochs} & \textbf{Time / Epoch} & \textbf{Total Time} \\
\midrule
QDrop (PTQ) & No & Yes & -- & -- & 0.2 h \\
LSQ (full QAT ref.) & Yes & Yes & 200 & 0.55 h & 110 h \\
ETBQ + QDrop & Yes & No (ETBQ), Yes (QDrop) & 80 & 0.38 h & 30.6 h \\
\bottomrule
\end{tabular}
}
\end{table}

ETBQ is designed as a tuning-before-quantization stage and therefore uses the training data to update FP weights before PTQ. 
Table~\ref{tbl:cost_comparison} compares the computational cost of ETBQ with representative PTQ and full-QAT pipelines using ResNet-18 on ImageNet. QDrop is the most efficient because it only performs post-training calibration and reconstruction on a fixed FP model, requiring about 0.2 hours in our setting. LSQ is included as a full-QAT reference: it jointly updates FP weights and quantization parameters over a 200-epoch training schedule, leading to a total cost of 110 hours. ETBQ lies between these two regimes: it updates FP weights during an 80-epoch pre-conditioning stage but does not optimize quantization parameters during pre-conditioning, and the reported 30.6 hours already includes the subsequent QDrop calibration. This comparison clarifies the cost position of ETBQ relative to calibration-only PTQ and standard full QAT; it is not intended as an equal-budget QAT accuracy comparison.

\section{Conclusion, Impact and Limitation}

This paper revisited low-bit PTQ from a pre-quantization robustness perspective. We argued that low-bit PTQ degradation is not only a post-hoc calibration problem: when the FP model is sensitive to quantization-scale perturbations, even calibrated weight and activation errors can lead to substantial accuracy loss. To address this issue, we proposed ETBQ, a tuning-before-quantization framework that pre-conditions the FP model before PTQ. ETBQ combines drift-controlled Weight Quantization Noise (WQN), salience-aware Activation Quantization Noise (AQN), and Stochastic Weight Averaging (SWA) to improve robustness to weight- and activation-side perturbations while remaining compatible with reconstruction-based PTQ backends. Experiments on CIFAR-100, Tiny-ImageNet, ImageNet, and Cityscapes show consistent improvements over direct PTQ baselines, especially under challenging settings such as W2A4. 

This work puts forward the core insight that the quantization tolerance of the FP model itself imposes a fundamental upper bound on final PTQ accuracy. This perspective shifts the research boundary from minimizing quantization error after the fact to pre-conditioning the FP model to be quantization-friendly beforehand, opening up a complementary upstream direction for the PTQ research community.

The current experiments mainly verify compatibility with reconstruction-based PTQ backends, especially QDrop and QEP; extending the evaluation to data-free PTQ and broader rounding-optimization pipelines is left for future work. In addition, WQN and AQN model quantization errors with Gaussian approximations. This is effective as a first-order proxy in our experiments, but the approximation becomes less accurate under extremely low bit-widths such as W2A2, where the observed gains are smaller. Finally, the present evidence supports a correlation between flatter FP loss basins and improved quantization behavior for SGD-trained models; extending ETBQ to Adam-trained models remains an important direction for future work.

\appendices
\section{Quantization-Error Estimation Details}
\label{app:quant_error_modeling}

\subsection{Empirical Justification for Gaussian Modeling}
\label{app:error_distribution}

\begin{figure}[h]
    \centering
    \includegraphics[width=1.0\linewidth]{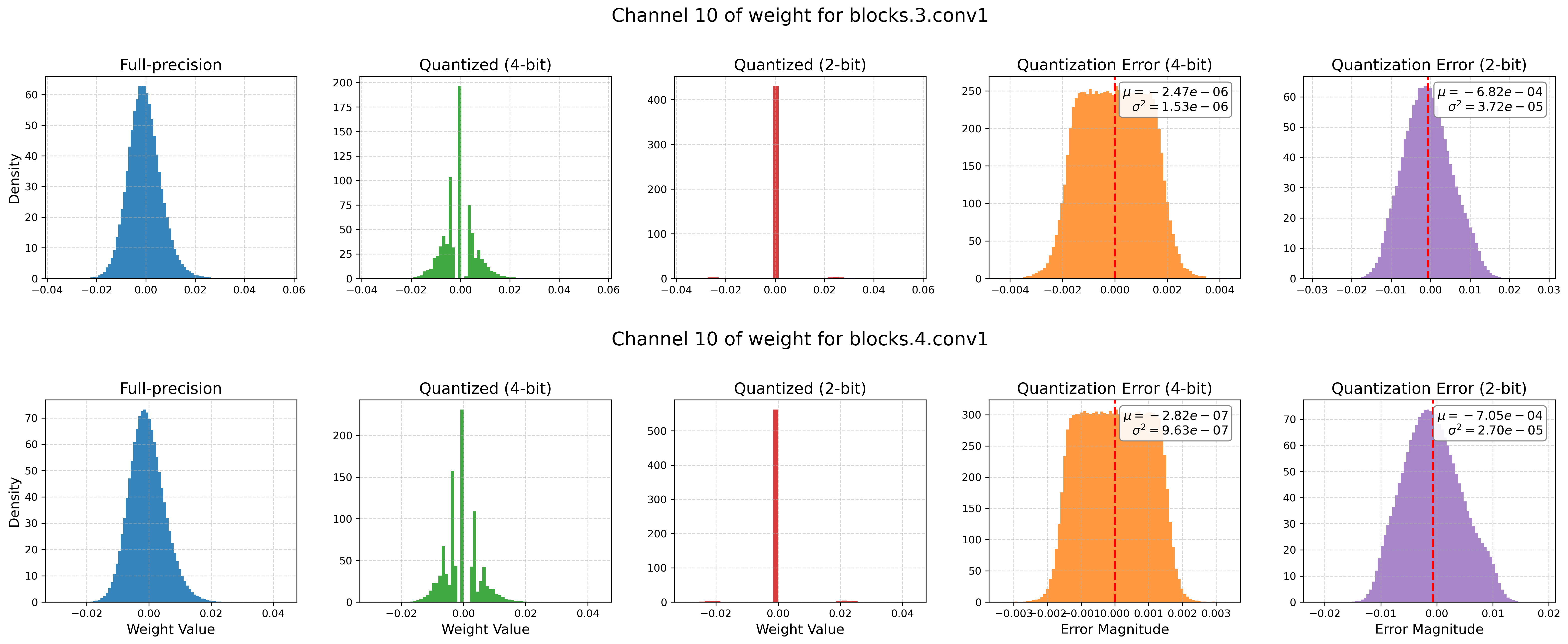}
    \caption{Representative weight and quantization-error distributions from different ResNet-18 blocks on CIFAR-100.}
    \label{fig:app_error_distribution}
\end{figure}

Figure~\ref{fig:app_error_distribution} shows representative weight and quantization-error distributions from different ResNet-18 blocks on CIFAR-100. The errors are generally centered around small values and exhibit a unimodal shape, especially under 4-bit quantization, supporting the Gaussian first-order approximation used in WQN. Under 2-bit quantization, the distributions become wider and less Gaussian-like, which partly explains why extremely low-bit settings remain more challenging.

\begin{figure}[h]
    \centering
    \includegraphics[width=1.0\linewidth]{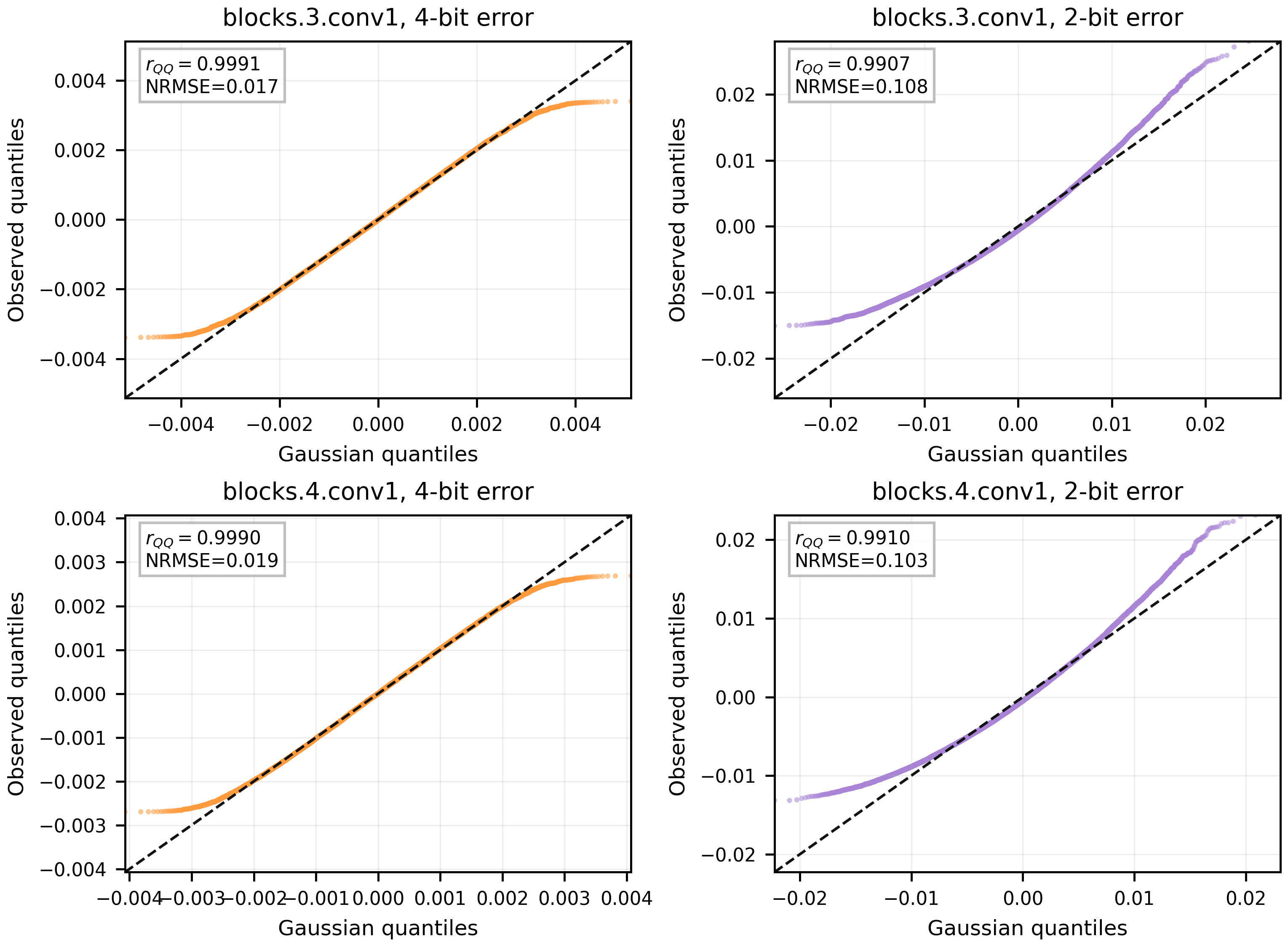}
    \caption{Gaussian Q-Q diagnostics for representative channel-wise weight quantization errors in ResNet-18 on CIFAR-100. The dashed line denotes the Gaussian reference with matched empirical mean and variance. Points closer to the diagonal indicate better Gaussian approximation in the corresponding quantile region.}
    \label{fig:qq_error_distribution}
\end{figure}

Figure~\ref{fig:qq_error_distribution} provides Gaussian Q-Q diagnostics for representative channel-wise weight quantization errors. The Q-Q plots compare empirical error quantiles with the quantiles of a moment-matched Gaussian distribution using the same mean and variance. To quantify the approximation quality, we report the Q-Q correlation $r_{\mathrm{QQ}}$ and the normalized Q-Q root mean squared error $\mathrm{NRMSE}_{\mathrm{QQ}}$, computed over the central $1\%$--$99\%$ quantile range. In our analysis, a high $r_{\mathrm{QQ}}$ and a small $\mathrm{NRMSE}_{\mathrm{QQ}}$ indicate that the main probability mass can be reasonably approximated by the Gaussian surrogate. Empirically, the 4-bit errors achieve $r_{\mathrm{QQ}}\approx0.999$ and $\mathrm{NRMSE}_{\mathrm{QQ}}<0.02$, showing a close match in the central region. Under 2-bit quantization, $r_{\mathrm{QQ}}$ remains above $0.99$, but $\mathrm{NRMSE}_{\mathrm{QQ}}$ increases to about $0.10$, indicating more visible deviation from Gaussianity. These deviations are expected because uniform quantization induces bounded and discretization-dependent errors. Therefore, WQN uses the Gaussian model as a tractable moment-matched surrogate rather than a strict distributional assumption.

\subsection{Channel-wise Estimation of WQN Statistics}
\label{app:wqn_statistics}

For the weight quantization error in Eq.~\eqref{eq:wqn_error}, WQN estimates statistics at the same output-channel granularity as the channel-wise weight quantizer. For $\boldsymbol{W}\in\mathbb{R}^{C_{\mathrm{out}}\times C_{\mathrm{in}}\times K_H\times K_W}$, the $i$-th output-channel error $\boldsymbol{E}_{\boldsymbol{W}_i}$ contains $N_i=C_{\mathrm{in}}K_HK_W$ elements, and
\begin{equation}
\label{eqt:model_weight_error}
\begin{aligned}
\mu_{\boldsymbol{W}_i} &= \frac{1}{N_i}\sum_{j,h,k}\boldsymbol{E}_{\boldsymbol{W}_i,j,h,k},\\
\sigma^2_{\boldsymbol{W}_i} &= \frac{1}{N_i}\sum_{j,h,k}\left(\boldsymbol{E}_{\boldsymbol{W}_i,j,h,k}-\mu_{\boldsymbol{W}_i}\right)^2.
\end{aligned}
\end{equation}
This yields $\boldsymbol{\mu}_{\boldsymbol{W}_i}\in\mathbb{R}^{C_{\mathrm{out}}}$ and $\boldsymbol{\sigma}_{\boldsymbol{W}_i}^{2}\in\mathbb{R}^{C_{\mathrm{out}}}$, with covariance $\boldsymbol{\Sigma}_{\boldsymbol{W}_i}=\mathrm{diag}(\boldsymbol{\sigma}_{\boldsymbol{W}_i}^{2})$. Using the output-channel granularity avoids mixing channels whose quantization ranges differ.

\FloatBarrier

\subsection{EMA-based Estimation of AQN Statistics}
\label{app:aqn_statistics}

Given the activation quantization error in Eq.~\eqref{eq:aqn_activation_error} and the per-tensor Gaussian model in Eq.~\eqref{eq:aqn_distribution}, AQN maintains one scalar mean and variance for each activation quantizer. The estimates are updated across epochs by EMA:
\begin{equation}
\label{eqt:model_activation_error}
\begin{aligned}
\mu_{\boldsymbol{a}_l}^{(e)} &=
\beta \mu_{\boldsymbol{a}_l}^{(e-1)}
+(1-\beta)\,\mathbb{E}_{\boldsymbol{a}_l\in\mathscr{D}_c}
[\hat{\boldsymbol{a}}_l-\boldsymbol{a}_l],\\
(\sigma_{\boldsymbol{a}_l}^{2})^{(e)} &=
\beta(\sigma_{\boldsymbol{a}_l}^{2})^{(e-1)}
+(1-\beta)\,\operatorname{Var}_{\boldsymbol{a}_l\in\mathscr{D}_c}
(\hat{\boldsymbol{a}}_l-\boldsymbol{a}_l).
\end{aligned}
\end{equation}
Here, $\beta=0.9$ in all experiments. The expectation and variance are computed over all batch, channel, and spatial elements collected from the calibration subset.

\subsection{Salience-Aware Mask: Implementation Notes}
\label{app:aqn_mask_detail}

The salience-aware mask is defined in Eqs.~\eqref{eqt:channel_salience} and~\eqref{eq:aqn_mask_probability}. In implementation, the channel salience is normalized with the following numerically stable softmax:
\begin{equation}
\label{eqt:app_salience_softmax}
S_{b,c}=
\frac{\exp(I_{b,c}-\max_{c'} I_{b,c'})}
{\sum_{j=1}^{C}\exp(I_{b,j}-\max_{c'} I_{b,c'})}.
\end{equation}
The normalized score $S_{b,c}$ is then rescaled, clipped, and sampled following Eq.~\eqref{eq:aqn_mask_probability}. The sampled mask $\boldsymbol{M}\in\{0,1\}^{B\times C}$ is broadcast along the spatial dimensions before being applied to the activation perturbation.

\section{Implementation Details for ETBQ Pre-conditioning}
\label{app:hyperparameter_settings}

To make the experimental protocol reproducible, Table~\ref{tab:etbq_hyperparams}
summarizes the main hyperparameters used for ETBQ pre-conditioning across all
datasets. The calibration subset is used to estimate WQN/AQN error statistics
and to perform downstream PTQ calibration, while FP weights are updated using
the corresponding training set.

\begin{table*}[t]
\centering
\caption{Hyperparameter settings for ETBQ pre-conditioning across datasets.}
\label{tab:etbq_hyperparams}
\resizebox{\textwidth}{!}{
\begin{tabular}{lcccc}
\toprule
\textbf{Setting} & \textbf{CIFAR-100} & \textbf{Tiny-ImageNet} & \textbf{ImageNet} & \textbf{Cityscapes} \\
\midrule
Task & Classification & Classification & Classification & Segmentation \\
Architecture(s) & ResNet-18/50, MobileNet-V1/V2 & ResNet-18, GhostNet & ResNet-18/50, MobileNet-V1/V2 & U-Net \\
Calibration subset & 100 images & 200 images & 1024 images & 128 images \\
ETBQ training data & Full training set & Full training set & Full training set & Full training set \\
Pre-conditioning epochs & 120 & 120 & 80 & 500 \\
Training schedule & ETBQ + SWA & ETBQ + SWA & ETBQ + SWA & 300 clean + 100 ETBQ + 100 SWA \\
Batch size & 256 & 128 & 128 & 4 \\
Optimizer & SGD & SGD & SGD & SGD \\
Initial learning rate & 0.015 & 0.001 & $2\times10^{-5}$ & 0.01 \\
Momentum & 0.9 & 0.9 & 0.9 & 0.9 \\
Weight decay & $5\times10^{-4}$ & $1\times10^{-3}$ & $1\times10^{-4}$ & $5\times10^{-4}$ \\
Label smoothing / loss & 0.1 & 0.1 & 0.0 & Cross-entropy \\
Error warmup epochs & 20 & 20 & 3 & 10 \\
Maximum error intensity $\lambda_{\max}$ & 1.0 & 0.8 & 0.05 & 0.8 \\
AQN EMA momentum $\beta$ & 0.9 & 0.9 & 0.9 & 0.9 \\
Downstream PTQ backend & QDrop/QEP & QDrop & QDrop/QEP & QDrop \\
\bottomrule
\end{tabular}
}
\end{table*}

\section{Additional Studies}
\label{app:additional_ablation}

This section provides mechanism-level ablations for two internal design choices of ETBQ: temporal differencing in WQN and salience-aware masking in AQN. The main paper reports component-level ablations of WQN, AQN, and SWA.

\subsection{Effect of Temporal Differencing in WQN}
\label{app:wqn_temporal_differencing}

\begin{table}[h]
\centering
\caption{Effect of weight quantization error injection mechanisms in WQN on ResNet-18/CIFAR-100 under W2A4 quantization.}
\label{tab:app_ablation_differential}
\begin{tabular}{lccc}
\toprule
\textbf{Injection Method} & \textbf{FP32 Acc. (\%)} & \textbf{W2A4 Acc. (\%)} & \textbf{Drop} \\
\midrule
Baseline          & 79.40 & 76.47 & -2.93 \\
Naive Additive               & 79.48 & 77.01 & -2.38 \\
\textbf{Differential (Ours)} & \textbf{79.56} & \textbf{77.27} & \textbf{-2.34} \\
\bottomrule
\end{tabular}
\end{table}

Table~\ref{tab:app_ablation_differential} compares different weight quantization error injection strategies. Naive additive quantization error improves W2A4 accuracy from 76.47\% to 77.01\%, showing that weight perturbation exposure is beneficial. However, directly adding independently sampled perturbations can introduce drift when the empirical quantization-error distribution is not zero-mean. Differential WQN instead uses the differenced perturbation $\boldsymbol{P}_{\boldsymbol{W}_t}=\boldsymbol{\delta}_{\boldsymbol{W}_t}-\boldsymbol{\delta}_{\boldsymbol{W}_{t-1}}$, which controls trajectory-averaged drift while preserving quantization-aligned perturbation structure. It further improves W2A4 accuracy to 77.27\%, suggesting that WQN benefits from drift-controlled quantization-error injection rather than arbitrary random perturbation.

\subsection{Effect of Salience-Aware Masking in AQN}
\label{app:aqn_salience_masking}

\begin{table}[h]
\centering
\caption{Effect of activation masking strategies in AQN on ResNet-18/CIFAR-100 under W2A4 quantization.}
\label{tab:app_ablation_masking}
\begin{tabular}{lccc}
\toprule
\textbf{Masking Strategy} & \textbf{FP32 Acc. (\%)} & \textbf{W2A4 Acc. (\%)} & \textbf{Drop} \\
\midrule
Baseline (No Mask)              & 79.40 & 76.47 & -2.93 \\
Uniform Bernoulli               & 79.51 & 77.17 & \textbf{-2.34} \\
\textbf{Salience-Aware (Ours)}  & \textbf{79.81} & \textbf{77.41} & -2.40 \\
\bottomrule
\end{tabular}
\end{table}

Table~\ref{tab:app_ablation_masking} compares activation masking strategies. Uniform Bernoulli masking improves W2A4 accuracy from 76.47\% to 77.17\%, indicating that stochastic activation perturbation improves quantization robustness. Salience-aware masking further increases W2A4 accuracy to 77.41\%, showing that the perturbation location matters. By assigning higher injection probability to more salient channels, AQN provides a controlled activation perturbation mechanism rather than merely adding random activation quantization error.

\bibliographystyle{IEEEtran}
\bibliography{refs}

\end{document}